\documentclass[fleqn,a4paper,12pt]{article}

\usepackage[latin1]{inputenc} 
\usepackage{amsmath}          
                              
\usepackage{amssymb}  
\usepackage{stmaryrd} 
\usepackage{endnotes} 

\usepackage{setspace}  
\usepackage{newcent}

\makeatletter
\renewcommand{\section}{\@startsection
  {section}{1}{0pt}{-\baselineskip}{.5\baselineskip}{\large\bfseries}}
\makeatother

\catcode`\@=11
\newcommand{\qbar}{\protect\@qb@mathpal}
\def\@qb@mathpal{\mathpalette\@qb@qbar}
\def\@qb@qbar#1#2{\let\@tempa#1\@qb@xqbar#2\@nil}
\newmuskip\@qb@muskip
\def\@qb@xqbar#1#2\@nil{
  \ifcat\noexpand#1A\relax
    \@qb@muskip4.4mu
    \count@`#1%
    \advance\count@-64\relax
    \ifcase\count@
      \or\or\or\@qb@CG\or\or
      \or\or\@qb@CG\or\or\or
      \or\or\or\or\or\@qb@CG
      \or\or\@qb@CG\or\or\or\@qb@TU
      \or\@qb@TU\or\@qb@TU\or\@qb@TU\or\or\@qb@TU
      \or\else\@qb@TU
    \fi
  \else
    \@qb@TU
  \fi
  \setbox\z@\hbox{$\m@th\@tempa#1#2$}
  \rlap{$
    \m@th\@tempa
    \mkern\@qb@muskip
    \overline{
      \phantom{\copy\z@}
      \mkern-\@qb@muskip
      \kern-.4pt
    }
  $}
  \box\z@
}
\def\@qb@CG{\@qb@muskip3mu}
\def\@qb@TU{\@qb@muskip1.6mu}
\catcode`\@=12 
\begin{document}

\bibliographystyle{chicago}
 \begin{center}
   \vspace{0.5cm} \Large \textbf{Language Heedless of Logic --
     Philosophy Mindful of What? Failures of Distributive and Absorption
     Laws}  \vspace{0.25cm}

\bigskip

\vspace{0.5cm}
\normalsize
Arthur Merin
\\Department of Philosophy
\\ University of Konstanz

\date{} \medskip
\end{center}

\normalsize 
\noindent 
\begin{quote}
  \small \textbf{Abstract}: Much of philosophical logic and all of
  philosophy of language make empirical claims about the vernacular 
  natural language.  They presume semantics under which `\textit{and}'
  and `\textit{or}' are related by the dually paired distributive and
  absorption laws.  However, at least one of each pair of laws fails
  in the vernacular.  `Implicature'-based auxiliary theories
  associated with the programme of H.P.\:Grice do not prove remedial.
  Conceivable alternatives that might replace the familiar logics as
  descriptive instruments are briefly noted: (i) substructural logics
  and (ii) meaning composition in linear algebras over the reals,
  occasionally constrained by norms of classical logic.  Alternative
  (ii) locates the problem in violations of one of the idempotent
  laws. Reasons for a lack of curiosity about elementary and easily
  testable implications of the received theory are considered.  The
  concept of `reflective equilibrium' is critically examined for its
  role in reconciling normative desiderata and descriptive commitments.
\vspace{3pt}


\normalsize\end{quote}

\vspace{0.1cm}

\linespread{1.13}\selectfont

\section*{Overview}

\noindent
This essay aims to show that there is something wrong with a most
popular and elegant hypothesis about the coordinative recursion of
meanings in \linebreak natural, vernacular languages. The hypothesis
entails that such recursion satisfies the lattice-theoretic laws of
classical and intuitionistic logic. The facts are otherwise, and not
altogether boringly so.\;  Section 1 recalls the tra\-dition of logical
semantics for natural, vernacular languages and some familiar ways of
addressing known wrinkles.  Section 2 presents armchair experiments
for testing some laws so far untested and finds them violated. Section
3 argues that the `Gricean' approach of pragmatic ampliation, when
explicated in a framework (e.g.  Gazdar 1979) that allows it to make
falsifiable predictions, does not save the phenomena for logic.
Section 4 asks what will save the phenomena.  It briefly considers,
but neither expounds nor tests substructural logics and a conceivable
semantics in linear spaces.  Section 5 returns to the normative and
descriptive commitments of applied philosophical logic and observes
that conflicts are sometimes dealt with by appeal to the notion of
`reflective equilibrium'.  The line taken is against doing so.
Section 6 concludes with an outlook. The Appendix is a heuristic
towards meaning composition in ordered vector spaces on which
`\textit{and}' and `\textit{or}' denote various instances of linear
combination.

\section{True religion}

If there is anything which holds together the current mainstream of
\linebreak analytic philosophy, it is one composite assumption. The
assumption is that logic provides a basic framework both for norms of
right reasoning and for descriptive meaning theories of our natural,
everyday vernacular.%
\footnote{\small `Logic' means by default classical or intuitionistc
  logic.  `Meaning theory' refers to the implicit theory by which
  speakers of a language attach meanings to the phonological and
  syntactic objects that they produce and perceive. The `theory of
  meaning' aims to make it explicit. I think the late Michael Dummett
  made this nonce terminological distinction at one time.}
Philosophers reason in the vernacular most of the time, so the second
part of the assumption is a prerequisite for daily practice conforming
to the first. The twofold idea is all but taken for granted in the
typical introductory courses in logic and philosophy of language. It
is not decisively repudiated at research level. 

The basic compositional component of meaning, on this view, is given
by logical syntax and by semantics in truth or verification
conditions.  This component is held to come in a wrapping of
conversational pragmatics and perhaps other, purely conventional
speech act paraphernalia and assertibility requirements.  The wrapping
serves to take up the slack between the predictions of logic and the
philosopher's phenomenological data.  These data are in large parts
spontaneous native speaker intuitions on acceptability and paraphrase
of word strings, extended by judgments of coherence for sets of
strings.  When the strings should be grammatical sentences by rules of
syntax and the slack is in places where the logical operation meanings
appear impugned, the extra-logical accretions must square accounts.

The paradigm for this descriptive strategy is set by Frege.  For
sentence schema `\textit{A but B}', he claims truth-conditions $A\land
B$ and the intimation of a contrast between $B$ and what should be
expected in view of the foregoing.%
\footnote{\small Frege (1879:\textsection 7).  I substitute the
  English equivalent for his German.  A fairly comprehensive theory of
  `\textit{but}' in the doxastically interpreted probability calculus
  is in Merin (1999). In the present text, logical symbols have their
  classical interpretation. I use them indifferently to denote
  truth-functional connectives and concomitant operations of a boolean
  algebra of denotations, be it of sets of states of affairs, elements
  of a Lindenbaum algebra, or \textit{sui generis}. This is how many
  classic texts on boolean algebras have proceeded. If you prefer,
  imagine lattice operator symbols `$\sqcap$', `$\sqcup$',
  `$\sqsupset$', etc. In algebraic mode, I use the equality symbol
  `$=$' where logical syntax would have a biconditional.  Classical,
  material implication is denoted by the horseshoe `$\supset$'.
  `$\textsc{xor}$' designates exclusive disjunction.  }
For the warranted felicitous assertibility of `\textit{If A, then C}
he gives three conditions: ($\alpha$) The material implication,
$A\supset C$, must be known to be true; ($\beta$) the truth values of
$A$ and of $C$ must be unknown; ($\gamma$) there must be some
connection of cause or necessitation between $A$ and $C$.%
\footnote{\small'\label{FREGE-IF} Op.cit.\:\textsection 5. Frege is
  not widely known for this.  In \textsection 12, he explicates
  ($\gamma$) by instantiation of a lawlike generalization
  \mbox{$\forall x[Px\supset Qx]$} (`for any individual $x$: $x$
  having property $P$ implies $x$ having property $Q$'). Set $A = Pa$
  and $C = Qa$ for some $a$.  However, this is not the only
  conceivable approach to the elusive requirement ($\gamma$).
  \textbf{Theorem}: When ($\alpha$) and ($\beta$) are explicated in
  the doxastically interpreted probability calculus as ($\alpha\pi$)
  $P(A\supset C)=1$ and ($\beta\pi$) $0 < P(A), P(C) < 1$, they
  jointly entail ($\gamma\pi$) $P(C|A) > P(C)$, which explicates
  positive evidential relevance of $A$ to $C$.  \linebreak
  \textit{Proof}: easy exercise; ($\beta\pi$) can be weakened to
  ($\delta\pi$) $0\neq P(A) \,\land\, P(C)\neq 1$, since $\alpha\pi,
  \delta\pi \vdash \beta\pi$.  $\Box$\; The probabilistic doctrine of
  `\textit{but}' presented in Merin (1999) extends to predicate
  languages for inductive reasoning. Like the application of the
  theorem, this work presupposes a classical logical skeleton for the
  vernacular.}

With reference to `\textit{A or B}', Ernst Schr\"oder (1890:134f.)
establishes a sub-paradigm of meaning supplements which are
rationalized by appeal to desiderata of cooperative conversational
conduct.\label{SCHROEDER}%
\footnote{\small Schr\"oder is best known for the first, if somewhat
  staggeredly presented axiomatization of boolean algebras and for
  having lent his name to the Schr\"oder-Bernstein Equivalence
  Theorem, of which (as Felgner 2002:587 recalls) he had offered a
  first, albeit defective proof sketch.}
The usual intimation of `\textit{A or B}' is that the assertor does
not know which of the disjuncts is true. The rationale is found by
\textit{reductio}. The truthful speaker must know that $A\lor B$ is
true. Had he known which of $A$ or $B$ was true, we expect that he
would have affirmed a (i) ``more informative'' and (ii) ``shorter''
expression alternative, namely one of $A$ and $B$.  Since he did not,
he will not know, and if he did know all the same, we should feel that
we had been misled.%
\footnote{\small \label{TARSKI}Schr\"oder's argument is taken up in
  passing by Tarski (1946:\textsection 8), who uses \linebreak
  \textit{Algebra der Logik} extensively in 1930s/40s research work,
  in Quine (1950:\textsection 3), who also transposes to
  `\textit{if}', and finally, with higher profile, in Grice (1961).
  The Schr\"oder-Grice intimation also entails that the speaker does
  not know whether $A\land B$, the putative denotation of `\textit{A
    and B}', is true. A stronger intimation would be that he knows it
  to be false, i.e.~that $A$ and $B$ exclude one another. This
  intimation, which is often felt to be made, if for the most part
  vaguely so, is not entailed by the Schr\"oder-Grice assumptions.
  However, suppose the predominant vagueness of the mutual exclusion
  intuition is taken seriously as a datum, and not simply treated as
  noise in data collection or a reflex of unresolved ambiguity. And
  suppose a probabilistic doxology is again adopted. Then the Theorem
  of note\:\ref{FREGE-IF} has, by way of Frege's assertibility
  doctrine and his definition `$A\lor B$ iff $\neg A\supset B$', a
  pertinent \textbf{Corollary}: Felicitously asserted $A\lor B$ always
  has $A$ negatively relevant to $B$ (i.e.~ $P(B|A)<P(B)$) and, of
  course, vice versa. The special case of extreme negative relevance
  (when $P(AB)=0<P(A),P(B)<1$) will explicate unvague intuitions of
  disjointness.}
Under the programmatic rubric `Logic and Conversation' (Grice 1967),
this doubly rational enterprise has captured the imagination of the
analytic mainstream and of its linguistic derivatives.%
\footnote{\small The programme accomodates, as lexically ambiguous,
  words which have a logical rendering, but which also have
  occurrences that must \textit{a priori} refuse it.  Example:
  `\textit{and}' will denote arithmetical addition, `$+$', in
  `\textit{Two and two is four}'. --- Highly involved theories, be
  they deterministically ontological, probabilistic or otherwise
  plausibilistic in the sense of graded modality, have been and are
  being offered for conditionals, both indicative and subjunctive.
  Their logics are eminently non-classical, lacking notably
  monotonicity a.k.a.  `Weakening' or `Thinning' (i.e. $A$$>$$C$
  \,$\not\vdash$\, $AB$$>$$C$, see e.g.~Lewis 1973), but they presuppose a
  classical logic for all modal- and conditional-free fragments of the
  language.}
 Gone are the days when the later Wittgenstein and J.L. Austin could
 persuade sizeable philosophical constituencies to do without the
 assumption of a logical skeleton and muscle to our vernacular, or
 even to deny the assumption.

 Victorious logic comes with interpretations in truth conditions or
 warranted belief-conditions. In this form, the tenet that logic
 ---here understood as in logic primers or in a non-monotonic variant
 as in Lewis (1973)---supplies the basic meaning theory of the vernacular
 is a deep secular conviction of the analytic trade, at the very least
 of a very prominent faction of it.  The two most obvious reason for
 its hold on the imagination are two findings, one positive, one
 negative, by philosophers who are at ease with elementary logic. (A)
 They find overwhelming evidence for a high degree of compositionality
 in everyday language. (B) They find it hard to conceive of a matching
 meaning theory worth the name that is not, at the core, hard logic
 supplemented by pragmatic wrappings of varying softness.  But the
 tenet about logic also has the status of an article of faith.
 Analytic philosophers use the vernacular as a would-be universal
 language, like everyone else.  Unlike everyone else, however, they
 also want that which they hear and say to be intelligible by the gold
 standard of intelligibility.  Logic provides that standard. For a
 suitable choice of logic, classical logic for many of us, it provides
 the gold standard of rationality.

 The predicate `rational' is, of course, commendatory, indeed emotive.
 We notice at the latest when considering its contrary, `irrational'.
 Unless one is out to upset the bourgeois, rational is something one
 should wish to be. But one's language is a part of one's being that
 is criterial for the attribution of rationality, and a conviction
 that matches a wish in the manner of Hegel's dictum about the Real
 being Rational is wish-conforming.  What conformity to the wish 
 adds to the conviction is a potential for confirmation bias, a
 reduced willingness to test the conviction as assiduously as any
 other scientific hypothesis.  The tendency does not imply that our
 thinking has been wishful. Only a demonstration of the tenet's
 falsity could do that.  But it does imply an attitude characteristic
 of True Religion which is independent of the truth, falsity, or
 meaninglessness of that faith's world-descriptive claims.

Now, if philosophy tries to live up to its characteristic tradition of
self-reflection, logic as putatively descriptive of vernacular meaning
must be more than a matter of conviction-by-default let alone blind
faith. Logic will also be part of an empirical science, much as a
mathematical theory of gravitation and a theory of one's scientific
instruments are part of physics.  Such a science must be experimental
in one way or another, and indeed vernacular-describing science
\textit{is} experimental.  Philosophers have for long conducted
armchair experiments on what strings of words make sense or are
apparent non-sense. They intuit what sentences follow from what
sentences and whether pairs of sentences are equivalent in meaning.
These sentences may be found in print or made up on the spot.
Thus, one should have expected at least routine armchair testing of
the basic laws of the presumed theory. People had, after all, bothered
to test Newton's laws of motion and gravitation with clocks, balances,
measurement rods, and, if need be, the aid of vacuum pumps.

However, this humdrum expectation is wide of the mark. There is no
record of the critical experiments having been conducted.  One might
conclude that philosophers and linguists take the object of language
science to be less important than physicists have taken that of
physical science.  The conclusion would be consistent with their
simply not bothering to check. But in the cases to be examined, the
crucial experiments of first resort are so obvious and so inexpensive
to run that a slightly different hypothesis would be no less well
supported. The hypothesis would be that contemplators who are
independent-minded enough not to take easily testables for granted
have, without quite realizing it, adopted an attitude of studied
disregard.  This hypothesis motivates our section heading.  It also
motivates a bit of tedium to come. Experimentation, as
post-Aristotelian `natural philosophy' realized, is about closing
loopholes to false doctrine. This is also what cogent argument is about.

\section{Its well-kept little secret}

Here is the kind of armchair experiment which is never conducted in
the literature.%
\footnote{\small Two near-exceptions are known to me, from 1914 and
  1985. They are discussed in sections 4 and 5, respectively.}
The experiment consists of two parts. Part 1 might offer for
contemplation (psychologists would say: as an experimental stimulus)
this pair of suitably anodyne word strings:
\begin{tabbing}
  \hbox{}\quad\= (1a)\; \= Anna is affable\textbf{,}\; and Brenda 
is benevolent or Cindy is careful.\\
\> (1b) \> Anna is affable and Brenda is benevolent\textbf{,}
 \; or Anna is affable and \\
\>\> Cindy is careful.
\end{tabbing}
The typographic convention is that the bolded comma followed by an
extra space represents prosodic grouping. The auxiliary theoretical
presumption will be that grouping represents `scope', i.e.~ ordering
of semantic recursion. Thus in (1a), what `\textit{or}' stands for
will be presumed to be applied to form a compound before the denotation
of `\textit{and}' is applied to this compound and a second conjunct.
The canonical translation into mathematical bracketting sees (1a)
bracketted as `\textit{Anna is affable and \textup{(}Brenda is
  benevolent or Cindy is careful\textup{)}.}', and (1b) as
`\textit{\textup{(}Anna is affable and Brenda is benevolent\textup{)}
  or \textup{(}Anna is affable and Cindy is careful\textup{)}}'.

Instructions to contemplators are twofold. (I) Judge for each of (1a)
and (1b) whether it is intelligible or at any rate acceptable as a
well-formed utterance of English! (II) Judge whether or not (1a) and
(1b) are equivalent in meaning!  Readers can now perform the
experiment inexpensively in the double role of experimental subject
and observer.  The prediction is that (1a) and (1b) are each found to
be well-formed and intelligible and to be equivalent in meaning: if
either one is to be judged true (or false) so is the other.
Affirmations for (II) would presumably entail affirmations for each
question of (I).

Part 2 of the experiment would repeat the procedure upon having
(1a) and (1b) replaced with examples (2a) and (2b):.
\begin{tabbing}
  \hbox{}\quad\= (2a)\; \= Anna is affable\textbf{,}\; or Brenda 
is benevolent and Cindy is careful.\\
\> (2b) \> Anna is affable or Brenda is benevolent\textbf{,}
 \; and Anna is affable or \\
\>\> Cindy is careful.
\end{tabbing}
I predict: (2a) will be found acceptable and intelligible. (2b) will
be found odd -- in robuster language, `weird' -- or indeed
unacceptable as a felicitous utterance and will quite possibly be
found unintelligible in virtue of this ill-formedness.  (2a) and (2b)
will not be judged intuitively equivalent in meaning. We can leave
open whether or not this is owed to the weirdness of (2b).  Replacing
(2b) by (2b$'$) `\textit{Anna is affable or Brenda is
  benevolent\textbf{,} \; and Cindy is careful or Anna is affable}',
will not in any significant way change the pattern of judgments.
Observe that the occurences of `\textit{and}' and `\textit{or}' are
all of the unexotic, sentence-conjoining, order-insensitive kind. They
ought to translate well into elementary logic, not as `\textit{and}'
fails to in `\textit{Kim and Sandy are a happy couple}', or
`\textit{It is possible to see Naples and die, but impossible to die
  and see Naples}'.

That said, the experimental paradigm is robust across `coordination
reduced' uses of the connectives. The reduced sentences are less
unwieldy, yet their synonymous re-expansion shows that the connectives
retain their unexotic, sentence-connecting properties. Thus, we find
the same pattern as above for pairs of sentence pairs whose second
pair (structurally akin to 2a,b) is
\begin{tabbing}
  \hbox{}\quad\= (3a)\; \= Kim is affable, or 
  \footnotesize she \normalsize is benevolent and careful.\\
  \> (3b) \> Kim is affable or benevolent, and 
\footnotesize she \normalsize is affable or careful.
\end{tabbing}
The small print for `\textit{she}' indicates de-stressing, which
ensures that `\textit{she}' refers anaphorically to Kim. Using the
optional pronoun here is a way of ensuring groupings as intended
before and thereby, one hopes, the associated scope relations of
`\textit{and}' and `\textit{or}'. A noticeable hiatus after the comma
can thus be dispensed with and the results confirm that the
unacceptability of (2b) is unlikely to be due to confusion about
groupings. The same response pattern as for (2) and (3) also attends
sentence coordination reduced into subject position. Here the optional
predicate occurrence printed in parentheses can be used as a grouping
device that makes reliance on prosody superfluous.
\begin{tabbing}
  \hbox{}\quad\= (4a)\; \= Anna (came)\textbf{,}\; or Brenda 
and Cindy came.\\
\> (4b) \> Anna or Brenda (came)\textbf{,}
 \; and Anna or Cindy came.
\end{tabbing}
The reduced analogues of (1a,b) will elicit the same doubly
affirmative judgments as the original. To see the import of these
findings, recall that our working sentential logics, among them most
prominently classical logic and intuitionistic logic (for which
`\textit{A or not A}' is not a tautology), validate the dual pair of
distributive laws:%
\footnote{\small There is no universal numbering convention for the three
  dual pairs of laws we shall consider. Some authors state first that
  law which has `$\land$' as the first or sole connective in its
  standard form, others opt for `$\lor$' first. For simple rhetorical
  effect, I shall order pairs so that the first-numbered of each pair
  corresponds to the (more) mellifluous English form.}
\begin{tabbing}\label{DIST} 
  \hbox{}\quad\= (Dis.1)\; \= 
 $A\land(B\lor C) = (A\land B)\lor (A\land C)$.\\
 \> (Dis.2) \>
 $A\lor(B\land C) = (A\lor B)\land (A\lor C)$. 
\end{tabbing}
Here `$=$' may be interpreted as logical equivalence \textit{qua} or
as algebraic identity..  (Dis.2) and (Dis.1) are interderivable in
lattices which generalize boolean algebra, formerly known as `the
algebra of logic'. In lattice symbolism, the relation schema `$X \le
Y$' stands for logical `$X$ entails $Y$', and `$X = Y$' thus stands
for reciprocal entailment.%
\footnote{\small Reminder: the Lindenbaum Algebra of a language $L$ of
  classical logic, whose elements are the equivalence classes of
  logically interderivable sentences of $L$, is a boolean algebra.  An
  arbitrary lattice (see briefly e.g.~ Mendelson 1970:\:Ch.\:5),
  unlike the boolean variety, need not have an operation corresponding
  to negation, and need not satisfy (Dis).  In lattice terminology,
  `$\land$' and `$\lor$' instantiate `meet', and `join', respectively.
  Let us very generously call `\textit{familiar}' any sentential logic
  whose algebra of sentence equivalence classes modulo
  interderivability is a lattice.}

The data from (2), (3), and (4) tell us that no logic validating
distributivity is \textit{prima facie} descriptively adequate, because
(Dis.2) fails to be validated by intuitions (i.e.~spontaneous native
speaker judgments) on acceptability and paraphrase.%
\footnote{\small \label{QUANTUM} In the very different descriptive domain of
  reconstructing how scientific measurements are combined,
  distributivity appears to fail for crucial instances in quantum
  mechanics, unlike in classical mechanics (Birkhoff and von Neumann
  1936). Measurement statements are identified with whole subspaces of
  a system-state vector space. The subspaces form a lattice, with
  $\land$ as intersection and $\lor$ as `linear span' (never mind the
  latter's exact definition). Under the canonical mapping of
  combining-operations to statement connectives, the combination law
  which fails on quantum physical grounds is not (Dis.2), but
  (Dis.1).}
Suppose failure \textit{prima facie} does persist \textit{secunda
  facie} after we have failed to come up with credible auxiliary
theories which save the phenomena for logic. Then we might conclude
that the logic of our vernacular language, \textit{as manifest in
  paraphrase and acceptability judgments}, is one whose algebra must
be a non-distributive lattice. But this conclusion is premature. All
lattices and all logics proposed for general-purpose, rational,
declarative argumentation satisfy the dual pair of Absorption Laws
which may, but need not, be seen as the special case $C;=A$ of the
distributive laws:
\begin{tabbing}
  \hbox{}\quad\= (Abs.1)\; \= $A\lor(A\land B) = A$.\\ 
\>(Abs.2)\> $A\land(A\lor B) = A$.
\end{tabbing}
The scientifically obvious move will now be to elicit judgments of
acceptability and paraphrase for corresponding candidate instances.
\begin{tabbing}
  \hbox{}\quad\= (5a)\; \= Anna is affable\textbf{,}\; or 
 Anna is affable and Brenda 
is benevolent.\\
\> (5b) \> Anna is affable.\\
\> (5c) \> Anna is affable\textbf{,}
 \; and Anna is affable or Brenda is benevolent.\\
\> (5c$'$) \> Anna is affable\textbf{,} \; and Brenda is benevolent 
or Anna is affable.
\end{tabbing}
We find that each of (5a) and (5b) is individually acceptable and
intelligible, but that the pair are not judged to be equivalent in
meaning. It takes considerable indoctrination -- in the noblest of
senses -- into norms or conventions of argument to be convinced that
an utterance of (5a) deductively commits the speaker or the believer
to no more and no less than (5b).%
\footnote{\small Ignore that (5a) commits us to Brenda's existence:
  recall example forms (3a,b).}  
But suppose that this indoctrination is sucessful or that our intuitor
is a natural born logician and will immediately spot that all the
speaker of (5a) can be nailed down to in adversarial dialogue is (5b).
Then the real trouble is yet to come..

Sentence (5c) will be judged weird or indeed unacceptable. So will its
variant (5c$'$), synonymous by intuitive and logical commutativity of
`\textit{or}'. This indicates that purely syntactic confusion with a
schema `(\textit{A and A}) \textit{or B}' cannot explain why (5c) is
bad. It follows for reasons apparent in (2b) that (5a) and (5c) will
not be judged intuitively equivalent.  This does not preclude that
\textit{secunda facie} construals of (5c) which make for
intelligibility in spite of weirdness will likewise fail to be judged
equivalent to each of (5b) and to (5a). Conjunction-reduced analogues
to (3) and (4) will follow the same pattern as (5) does.

Thus, English and similar languages fail to validate (Dis.2) and each
of (Abs.1) and (Abs.2), but in asymmetric ways. The candidate instance
of (Abs.1) has each side of the equivalence acceptable, but fails
equivalence; while (Abs.2) fails already due to unacceptability of its
longer side. Lattices and their associated logics obey a Duality
Principle: any valid equality in `meet' (`$\land$') and `join'
(`$\lor$') terms remains valid if each connector is replaced by the
other. Apparent violations of duality will already dispose us to
conclude that, if any one of `\textit{and}' and `\textit{or}', label
it $\sigma$, fails to denote its logical correlate, so will the other,
dub it $\tau$. 

This heuristic can be filled in.  Suppose, as is likely, that there
are no other familiar logical correlates available.  For
`\textit{and}' there is no such candidate in sight, and \textsc{xor}
won't do for `\textit{or}' (see note\:\ref{XOR}).  Then `\textit{X
  $\sigma$ Y}' would, if at all, denote a complex which cannot be the
input to any other familiar logical connective.  But sentences (1a)
and (1b) are both perfectly good and in at least one of them $\sigma$
supplies an input to $\tau$, schematically `$Z \tau (X \sigma Y)$'.
Thus, we have a domino effect: if one logical interpretation goes, the
others go too. (`Unfamiliarly' logical $\sigma$ that save the
phenomena are not in my present sight.) Next, we consider another dual
pair of laws in the context of a conceivable remedy. This will point
the finger at `\textit{and}' -- in elementary, old-fashioned
philosophical logic the least controversial of connectives -- as the
primary problem.

\section{Grice will not save}

`Grice saves' was how the late linguist, James\:D.\:McCawley, titled a
section in his comprehensive book from the heyday of
logico-linguistics (McCawley 1981). In view of the reverential prefix
`Grice taught us that ...'  which one is apt to meet in the philosophy
of language, his two-word description of Grice's role seems like
doubly fair comment.  So: will He save here?  Initial cause for
optimism arises with the Idempotent Laws, which hold for all lattices
and known logics of general-purpose declarative mode argumentation.
\begin{tabbing}
  \hbox{}\quad\= (Ide.1)\; \= 
 $A\lor A = A$.\\
 \> (Ide.2) \>
 $A\land A = A$. 
\end{tabbing}
An instance experimental setup for testing their validation would be 
given by contemplata  
\begin{tabbing}
  \hbox{}\quad\= (6a)\; \= Anna is affable or Anna is affable.\\
\> (6b) \> Anna is affable.\\
\> (6c) \> Anna is affable and Anna is affable.
\end{tabbing}
The considered judgment will presumably be that each of (6a) and (6c)
is odd, weird, or indeed unacceptable.%
\footnote{\small With (6c) worse. 
 If its badness feels like giving way to a
  construal in terms of two distinct occasions of showing affability,
  use `\textit{Anna is tall}', `\textit{Anna is Austrian}' or
  suchlike.}
The question of their \textit{intuitive} equivalence to (6b) may
remain unanswered, because one is puzzled by them. If just
one example is found acceptable, it will surely be (6a). For (6c),
charitable reconstrual will presumably be needed.  Now, among Grice's
mutually known rules of cooperative conversational conduct there is
one that he dubbed the `Maxim of Manner', which went `Be
perspicuous!'. Its most tangible specifying submaximim is `Be brief',
or, in Grice's own paraphrase, `Avoid unnecessary prolixity' (cp.
Schr\"oder's (ii), p.\:\pageref{SCHROEDER} above).  It seems reasonable to
see the bare oddness judgments which attend (6a) and (6c) as being
predicted by ($\alpha$) the shared presumption that speakers avoid
needless prolixity and by Grice's further presumption ($\beta$) that
no overriding communicative purpose would be served by violation of
the maxim. Each of (6a) and (6c) is considerably more verbose than its
putative logical equivalent (6b).  There is no apparent reason, say,
etiquette or a quest for \textit{gravitas}, why the longer form might
be preferred, at any rate not before irony or sarcasm exploit the
perceived oddity.

Suppose we are satisfied with binary (in)acceptability judgments. Then
Grice does save for (Ide.1) and (Ide.2). This will be no mean feat,
because (Ide.2) is the modern way of expressing what Boole (1854:49)
called the \linebreak`\mbox{fundamental} law of thought',
characteristic of the algebra of logic. The\linebreak thought behind
it, plainly stated, is: `\textit{Saying the same thing twice over does
  not increase its evidential value}'.  Its great competitor is what
psychologists call the Law of Effect, whose relevant instance is:
`People will believe anything if you repeat often enough what speaks
for it'. This makes Boole's law a cherishable intellectual good, and
makes it antipsychologistic in a most unmetaphysical of senses.  If
Brevity saves it for language, we are in business.

However, Brevity will not explain the badness of the right-hand side
[RHS] of (2b), i.e.~the violation of (Dis.2).  This is because the RHS
of (1b), which would instantiate the RHS of (Dis.1) is equally long,
yet fully acceptable.  Similarly, acceptable (5a), which would
instantiate the left-hand side [LHS] of (Abs.1) is no less prolix than
unacceptable (5c), the would-be instance of the LHS of (Abs.2). If the
import of these observations were to restrict Brevity's explanatory
ambit to (Ide), the Gricean enterprise could count itself lucky. But
it cannot. The fully acceptable and intelligible RHS of (2a) [putative
for the RHS of Dis.1] is noticeably longer than its putative logical
equivalent LHS.  Moreover, the fully acceptable and intelligible (5a)
[putative for the LHS of Dis.2] is overwhelmingly longer than its
putative logical equivalent (5b). These observations show conclusively
that Brevity affords no explanation at all. Its apparent success with
(Ide) is spurious coincidence. 

Appeal to Brevity is also apt to make us overlook the very different
ways in which schemata `\textit{A or A}' and `\textit{A and A}' are
odd.  Take `\textit{A and A}' with stative $A$, say `\textit{Kim is
  tall}'. Statives (the taxonomy of which the term is part goes back
to Aristotle and has well-known 20th century developments by Anthony
Kenny, Zeno Vendler, and David Dowty) do not allow an additive
construal as `\textit{Kim talks and $($Kim$)$ and talks}' would.  With
statives (and also with `achievements' e.g.~`\textit{Kim turned 90'}
and `accomplishments' e.g.~`\textit{Kim broke the rear window}')
`\textit{A and A}' is irremediably weird.  Any use of it will be
sharply derogatory or insulting of someone, by default the addressee.
`\textit{A or A}', by contrast, is much less grating to the mind's
ear. It can be used as a bantering presentation of Hobson's Choice in
act or fact.%
\footnote{\small Is that philosophy? Well, it's not incuriosity.}  `Be
brief' has the virtue of brevity as an explanans, but few others.

Will `Be informative', the first of Schr\"oder's desiderata, as
rephrased by Grice under the label `Quantity', save the phenomena?
Informativeness, too, has received an intelligible explication among
Griceans, namely Schr\"oder's, in terms of comparative logical
strength.%
\footnote{\small Grice had two other maxims besides `Manner' (`Be
  Brief', `Be perspicuous') and `Quantity'. Of those, `Quality' says
  `Be truthful and warranted', and it resembles G.E.\:Moore's and
  Max\:Black's idea that speakers, in Black's (1952) diction,
  represent themselves as knowing or believing what they assert.
  `Relation' or `Be relevant'\linebreak has no explication in Grice,
  nor in work beholden to his.  His own 1967 example is someone
  abruptly starting to talk of the weather to intimate that the prior
  topic is embarrassing. Merin (1999) first examines a moderately
  protestant would-be alternative to Grice. This purports to
  characterize relevance and make it predict, but inadvertently
  reduces it to Grice's Informativeness and Perspicuity. The proposal
  in Merin (1999) is for explicating relevance with J.M.\:Keynes,
  Carnap and others in probability theory. This is applied to explain
  data addressed by Griceans.  References to uses of probability in
  the present essay refer to this approach.}
$X$ is logically stronger than $Y$ if $X$ entails $Y$, but is not
entailed by it. Example: let $X=A$, $Y=A\lor B$. `\textit{Or}' is
Schr\"oder's 1890 and one of Grice's 1961 paradigm examples for
reasoning by informativeness to generate what Grice called a
`generalized conversational implicature' attaching to an expression
type. Since `\textit{or}' occurs in putative correlates of (Abs) and
(Dis), Informativeness is a candidate explanation once the fate of
implicatures is accounted for when `\textit{A or B}' occurs in a
complex. By contrast, the schema `\textit{X and Y}' of which
`\textit{A and A}' is an instance has no Gricean implicature apart
from speaker's knowledge of its truth, i.e.~that of its conjuncts.
There remains Grice's `Relevance'. With Tarski loc.cit. we should
demand under this rubric that $X$ and $Y$ concatenated by either
connective not be too conceptually disparate, as `3 is prime' and `The
weather is fine' are.  But this is evidently not our problem.

Let us begin with the simplest schemata. The badness of `\textit{A and
  A}' remains unexplained. `\textit{A or A}' might draw on Frege's
supplementation doctrine, translated mechanically from `\textit{if}'
(p.\:\pageref{FREGE-IF} above) to `\textit{or}' by way of the
classical logical equivalence $(X\lor Y) \equiv (\neg X\supset Y)$,
i.e.~`$X$ \textsc{or} $Y$' is true iff `\textsc{not} $X$
\linebreak\textsc{implies}' $Y$' is. If the assertor of `\textit{X or Y}'
conventionally intimates (i) knowledge that $X\lor Y$ is true and (ii)
ignorance about the truth value of disjuncts, then instantiating each
of $X$ and $Y$ to \textit{A} will generate an epistemic contradiction,
since $A\lor A \equiv A$.  To be sure, a mechanical intimation of
ignorance would be required for this, and there must not be a
precedence protocol by which one of (i) and (ii) pre-empts the other.

Gazdar's (1979) seminal algorithm for assigning these formulaic kinds
of implicature to arbitrarily complex sentences, $S$, has a precedence
protocol on very general grounds.  `Assertions' of one subclause, $W$,
of $S$ may conflict with `potential implicatures' of another
subclause, $Z$. These would be the implicatures generated by
stand-alone utterances of $Z$. The protocol gives assertions
precedence, as it must, and keeps conflicting potentialities
unrealized.  This happens without a fuss.  But perhaps stand-alone
(6a), which paradoxically unites the roles of $W$ and $Z$, will make a
fuss, even though by Schr\"oder-Grice inference no implicature and
hence no conflict could arise.

The schema `\textit{A or B}' is often taken to generate a further
Informativeness implicature, namely that the speaker knows $A\land B$
to be false. This `strong' implicature will not arise by
Grice-Schr\"oder reasoning alone. Gazdar generates it as another
conventional, derogable default; Soames (1979) does so casuistically.%
\footnote{\small \label{XOR}For Soames, it arises when the speaker can
  be presumed to know that `$A\land B$' is true if it is true, and to
  know that it is false if it is false.  This double presumption and
  Schr\"oder ignorance jointly entail that he knows `$A\land B$' to be
  false. `Strong' implicature is to explain why `\textit{A or B}' is
  often (mis)construed as $A\,\textsc{xor}\,B$. There is indeed good
  reason to avoid \textsc{xor}: \mbox{`\textit{A or B or C}'} would be
  true iff an odd number of disjuncts are. In Merin (1994:~Ch.~3)
  Gazdar's algorithm for implicature projection is modified to cover
  $n$-fold disjunction for $n>2$. These cases have rightly been noted
  by McCawley (1981) to be intractable by Grice's original doctrine.
  (The adequacy proof for the extension is by complete induction. In
  subsequent work I have extended Soames's algorithm to $n>2$. There
  are differences potentially reflected in prosody.) The probabilistic
  approach of note\:\ref{TARSKI} above could motivate Gazdar's unvague
  lexical default assumption by a relevance-compositional rationale
  for $P(AB)=0$: this condition guarantees that the relevance of
  $A\lor B$ to any $H$ is a convex combination of the relevances of
  $A$ and of $B$ (Merin 2006:Th.3).  }
Applied to (6a) it would instantiate to: `speaker knows that $A\land
A$ is false', which reduces to `speaker knows that $A$ is false'.
Epistemic and aletheic paradox by contradiction with assertoric
`speaker knows that $A$ is true' is again avoided by sensible
precedence of assertions. I conclude that the explanatory potential of
Informativeness for intuitions about (6a), which relate to
`\textit{or}' and (Ide.1), is uncertain, and for those about (6c),
which relate to `\textit{and}' and (Ide.2), nil.

A like pattern emerges on applying Informativeness to (Abs).
(5a) is as acceptable as (5b). (5a) at first sight intimates
speaker's ignorance of which of its disjunct propositions, $A$ and
$A\land B$, is true. There can be no such intimation in (5b). A
failure of intuitive equivalence, \textit{contra} (Abs.1), could be
put down to this difference. But granted the primacy of assertion
(`speaker knows the disjunction to be true') over implicature, the
speaker of (5a) cannot be ignorant about $A$, since (5a) has the truth
conditions of (5b). To make the putative explanation of felt
inequivalence work, people's interpretive parts of mind must fail to
realize that (5a) has the truth conditions of (5b). But this is to
pull the rug from under the Gricean enterprise. The failure of
semantic competence could not be explained away as one of poor
`performance' due to limited working memory.  Our example sentences
are short already and the patterns persist for two-word instances of
`\textit{A}' and `\textit{B}'.

Undaunted, the supplementarian might address the weirdness of (5c) as
follows. Its two conjunct propositions, $A$ and $A \lor B$, are each
asserted. By the first, the Quality-conforming speaker must know that
$A$ is true.  The second has $A$ as a disjunct and so intimates that
the speaker does not know whether $A$ is true. So there is a
\textit{prima facie} contradiction of intimations.  In (5a), by
contrast, it may have taken some reasoning -- too much for the naive
intuitor -- to recognize a contradiction. But again: no contradiction
can persist in (5c) under any conceivable implicature projection
scheme. All must prioritize assertoric commitments and so block the
ignorance implicature from arising. For a Schr\"oder-Gricean, it could
not even arise momentarily.  Hence, the explanation attempt is again
one of uncertain purchase.

(Dis) fares worse. Weird (2b) is a putative instance of the right-hand
side of law (Dis.2). It offers no foothold even for mere attempts to
explain its weirdness by contradictory potential 
implicatures.  Assertion of the schema \mbox{\textit{$($A or B\/$)$
    and $($A or C\/$)$}} must generate implicatures of ignorance
about the truth values of $A$, $B$, and $C$.  These implicatures are
jointly consistent with speaker's knowledge, by `Quality', of the
truth of the non-implicatural content.  Adding `strong' implicatures
from conjuncts `\textit{A or B}' and `\textit{A or C}', namely that
the speaker knows each of $A\land B$ and $A\land C$ to be false,
preserves consistency.  No prioritization is needed.  Hence, the
uncertain explanation for the oddity of (5c) could be no more than a
fluke.  To sum up: the findings in this section cannot allay fears
that, in respect of logic and implicature, Grice was misled, and was
apt to mislead a congregation which wanted to be led exactly where he
did in fact lead them to -- the place they were already at.

\section{What will?}\label{WHAT}

Two kinds of conceivable salvation are at issue. One kind would save
logic -- that is: some logic widely acceptable as a working logic of
scientific and likeminded argumentation%
\footnote{\small As distinct, for instance, from a logic with models
  in chemical process engineering or in architectural design or in
  pattern constructions traditionally effected by categorial, extended
  Chomsky phrase structure or Lindenmeyer grammars.  See
  p.\:\pageref{LOGICS} on a logic with such models, \textit{inter
    alia}.}
-- as a theory which describes the recursive skeleton of our
vernacular meaning theory.  The other kind would merely save the
theory of meaning from the sceptical conclusion that there is no
theory worth calling so that will reconstruct our naive practice. More
specifically: it would save it from the conclusion that there is no
such theory which is as mathematically intelligible as a logical
theory, and thus conveyable in the Sciences' unambiguous
\textit{lingua franca}.

I know of no auxiliary theory that will deliver salvation of the first
kind. The \textit{prima facie} most obvious candidates in the paradigm
known as `Gricean' were found wanting in section 3.  One might thus
try to preserve logical conservatism by replacing Schr\"oder's and
Grice's most interesting resource, Informativeness defined by logical
entailment, by something else. The obvious candidate for those
familiar with the philosophy of science and the tradition of logical
empiricism will be inductive, that is, measure-theoretically
explicated relevance `for' or `against' a contextually given thesis.
Evidence $E$ for a thesis proposition $H$ makes $H$ more probable when
it is updated on, evidence against makes it less probable. A
corresponding change in conditional probability conditionalizes the
update relation, most literally so when updates are by conditioning a
probability function.

Relevance thus defined in the probability calculus presupposes and in
this sense conservatively extends classical logic (Merin 1997, 1999).
There was evidence for the advisability of a move from entailment to
thesis-driven relevance from the outset.  O'Hair (1969) observed that
Grice's Informativeness cannot in fact explain his very own key 1961
example, namely that ($\alpha$) `\textit{It looks red to me}'
implicates ($\beta$) `The speaker is not certain that it is red'. For
($\alpha$) is not, as the Gricean construal of `Informativeness' would
have to assume, logically weaker than ($\gamma$) `\textit{It is red}'.
Neither statement entails the other.%
\footnote{\small \label{HAIR} I have not seen a Gricean reply or
  acknowledgement in print.} The story for ($\alpha$) cannot then be
the Gricean story of \,`\textit{or}'.

What could explain the intimation ($\beta$)?  Suppose a context of use
in which ($\gamma$) is a stronger argument for some $H$ at issue than
($\alpha$) is. i.e.~ suppose that the assumption of ($\gamma$) raises
our degree-of-belief in $H$ (our personal probability that $H$ is
true) more than assuming ($\alpha$) does. (Say: $H=$ `It's oxide of
mercury', or $H=$ `It's a Communist flag'.) Grant also that the
paradigm for our vernacular discourse situation is issue-based and
thus at least in parts competitive, just as classical rhetoric
assumed.  Then we can infer the intuited intimation, namely that the
speaker lacks warrant for ($\gamma$).

Comparative Relevance so explicated is unlike comparative
Informativeness, which is not directional to some $H$ and so is
non-partisan. In Grice's deductive world, $A$ is more informative than
$B$ iff $A\models B$ while $B\not\models A$; making allowances for
degenerate entailments by the contradiction, e.g.~ $0=1$. Suppose
relative informativeness is itself defined more generally in
measure-theoretic terms as uncertainty-reduction.  Then it will be the
expectation, a probability weighted sum, of relevances.  (This is a
standard way to interpret `relative entropy', the quantity which the
update scheme of conditioning and a salient generalization of it
minimize.)  The expectation operator, as always, binds and thus
`kills' a variable.  Here, in particular, it thereby kills issue-based
directionality.  Directionality goes with debate or, less nobly put,
with persuasion in line with a speaker's interests. It does not go
well with Grice's quiet transformation of eminently partisan classical
rhetoric (whose theory of tropes harbours the inferencing principle of
implicatural indirection) into a pragmatics of cooperative, efficient
and, for theoretical purposes, disinterested information transmission.

Suppose our pragmatics were to be such.%
\footnote{\small  Modulo an account of
how the vernacular's compositional meaning engages the classical logic
of proposition spaces on which probabilities are defined, I think it
\textit{is} such.} 
Suppose it thus extended to engage `\textit{or}'. Then I would still
not see a conservative solution for all three dual pairs of problems.
Here, briefly, is a summary of why not.%
\footnote{\small Readers who use probability theory a lot will be on
  familiar ground when it comes to the basic tool. Others might be
  content to note that this approach has been tried. In the current
  state of discussion, it could itself be considered somewhat
  \textit{avant garde}, but for our present problem it would be, I
  think, another instance of rearguard action.}
Re (Ide.2): A probabilistically explicated Relevance requirement, dub
it `R', on `\textit{A and B}' could be that the amount of its
evidential relevance in favour of some logically independent
proposition $H$ at issue be construable as both non-nil and additive
by default. Specifically, additivity should be satisfiable under some
probability assignments and for a widely preferred relevance measure
such as the log-likelihood ratio (Merin 1999).%
\footnote{\small The measure's
aficionados include C.S. Peirce, D.  Wrinch and H.  Jeffreys, A.M.
Turing, and most prominently I.J.  Good.} 
`R' would be unsatisfiable for $B=A$ as in (6c). Why assume `R'? For
one, because a probability condition guaranteeing such additivity,
namely independence conditional on each of $H$ and $\neg H$, entails,
for $A$ and $B$ that are each positive to $H$ short of making it
certain, an ordering by increasing relevance: $A\lor B \prec X \prec
A\land B$, where $X$ is either of $A$ and $B$ when they are
equi-relevant, else the more positive (ibid.:Th.6).
Try:\:`Candidate\:1 has convictions for tax evasion or mail fraud'.
However, logic and `R' will not suffice to explain problems with (Abs)
and (Dis). Re (Abs.2): `R' would explain the badness of (5c) if $A\lor
B$ having zero relevance, e.g.~ with $A$ positive and $B$ suitably
negative to $H$, could be ruled out.  But I don't see how.  Re
(Dis.2): I do not see why bad (2b) could not satisfy `R', and indeed
consistently so with good (2a).

\label{KOENIG} There is one rather different, obscurely sited
near-proposal to report from Julius K\"onig (1914:75n1), which is also
the closest that the literature I know of has come to describing the
strange failure of (Dis.2).  K\"onig's stated aim, late in life, was
to found logic on a phenomenology of `undeniable experiences', or
`self-evidence'. He remarks -- no doubt with verbalized examples in
mind, but not giving any -- that (Dis.2) is not phenomenologically
`self-evident', whereas (Dis.1) is.  His explanation is (i) that
`\textit{or}' ambiguously denotes inclusive ($\lor$) and exclusive
(\textsc{xor}) disjunction and (ii) that the `self-evi\-dent' among the
laws of logic remain valid when `\textsc{xor}' replaces `$\lor$'.  It
is easily checked by truth-tables that (Dis.1) remains valid, while
(Dis.2) doesn't.  But if this hypothesis had been intended to explain
language phenomenology, it would fail to explain why (2b) is
unacceptable and not simply judged inequivalent to (2a).  The theory
would also falsely predict as being intuitively `self-evident' the
equivalence of (5a) and (5c), which would be implied by (Abs.1) and
(Abs.2), and it would leave unexplained the weirdness of (5c).
K\"onig indeed never mentions (Abs) among the laws of logic and almost
as an afterthought he introduces (Ide), which leaves him balancing in
precarious equilibrium on the fence betweeen psychology and either
sociology or ethics.  He notes that (Ide.2) states ``how we think
`\textit{A and A}' or, more exactly, how we \textsl{intend or decide}
to think (\textit{denken \textsl{wollen}}) \mbox{`\textit{A and A}'}''
(op.cit. 76).  The same is to be said about `\textit{A or A}'. (But
note: this is a contradiction-in-terms for \textsc{xor}.)%
\footnote{\small K\"onig is the father of K\"onig's Theorem and, as it
  were, the grandfather of K\"onig's Lemma. His posthumous book, seen
  through the press by his son, Denes K\"onig, also contains (then)
  advanced thoughts on set theory.  I chanced across it long after
  observing the facts of section 2. Perhaps there is a connection
  between K\"onig's phenomenological concerns and his tenet that some
  sets cannot be well-ordered, which he retained after two famous
  failed attempts to prove it for the continuum of real numbers.
  \linebreak Zermelo had followed each attempt with a proof to the
  contrary, namely that any set can be well-ordered. To do so, Zermelo
  had assumed the (now known-to-be equivalent) Axiom of Choice,
  insisting (van Heijenoort 1967:\:187) that the Axiom was
  ``self-evident''.}

I cannot, of course, rule out that a remedial auxiliary doctrine might
yet be found, either utilizing instruments inspired by the Gricean
enterprise or others. But a proponent of scepticism about
compositional logical semantics need not presently rule out such an
eventuality. By the evidential conventions of science and thus, I take
it, of philosophy, the burden of proof now rests with the proponent of
a logical skeleton conservatively supplemented by credible
conventional or `conversational' auxiliaries.

It may be objected that the skeleton is nowhere as rigid as I have
implied. Is there not consensus that `\textit{if}' needs a construal
in non-classical formalisms, e.g.~ for counterfactuals or when
negated?  Quite so, but the non-classical theories of `\textit{if}'
which seriously aim both to engage the vernacular and to retain
compositionality have each of `\textit{and}', `\textit{or}' and
`\textit{not}' retain their familiar logical meanings; see Lewis
(1973) as a representative of the field.%
\footnote{\small Adams (1975), who has a probabilistic theory fully
  based on belief or assertibility conditions, does introduce an
  assertoric `quasi-conjunction' and related disjunction with
  non-classical properties. However, these operations are subject to
  severe constraints on compositionality on pain of predicting very
  counterintuitive inferences.}  
If two of those go, the modernized logical skeleton will come apart.%

\label{LOGICS} Logic here means any logic validating the lattice laws.
In recent decades, logics have been discovered or developed which do
not validate all, or for that matter any of them. In the
proof-theoretic perspective usual of, and always initial to, their
treatment -- giving a highly general semantics for them is a tricky
task -- these logics fail to validate one or more of the `structural
rules' (see e.g.~Gentzen 1934) of traditional logics, among which are
some which are correlates of lattice laws. Accordingly, these
logics are referred to as \textit{substructural} logics (see e.g.
Paoli 2002, Restall 2000 for background). 

Some substructural logics notably do not validate (Ide.2), whose
proof-theoretic, structural rule correlate is `Contraction'. Among
these logics there is Classical Linear Logic [LL] with Exponentials
(Girard 1987) which has two conjunctions and two disjunctions. This
logic is one of `limited resources', because an object used in a proof
step, say by application of a Modus Ponens type rule, is used up, and
no longer available for another proof step. Indeed, one of the
earliest substructural logics, now known as the Lambek Calculus
(Lambek 1958), had its first application in modelling the parse or
syntactic production of a sentence as a proof, the objects of which
were syntactic constituent types. In such logics (Ide) or its
proof-theoretic correlate `Contraction' will typically fail
\textit{ceteris paribus}. However, Linear and similar logics can, as
it were, switch on (Ide) by use of its `exponential' operators.  The
exponential turns the formula $A$ from a scarce resource, whose single
syntactic occurrence is used up when used in inference, into an
abundant good, somewhat like a dish from the all-you-can-eat buffet.
With such devices, LL embeds classical logic. It also has a connection
to linear algebra, which was pointed out early on by Yves Lafont of
the ENS Paris and runs deeper than just the affordance of a
non-idempotent conjunction.  Semantics proposed for LL are very far\linebreak
from explicating truth-as-correspondence conditions and the most
intuitive of them have been in terms of strictly competitive games, as
presaged in \mbox{Lafont's} work.

I have not got LL to generate intuitive meanings for a usefully-sized
fragment of English.%
\footnote{\small I was first apprised of LL in 1988 (by Martin Hyland
  of Cambridge University) and thereupon started trying.}  
Neither have I managed to do so in a revealing way even for minuscule
fragments, say, for uses of vernacular `\textit{or}'. Example: Girard
(1990) brought to popular attention an appetizing menu-choice
illustration of the LL pair of `disjunction' connectives, with credit
to Lafont who invented it.  The two connectives are each paraphrased
by vernacular`\textit{or}', but one of them is apt for `free choice'
uses of `\textit{or}' (you pick one of soup or salad), the other for
uses involving pot-luck ignorance which will correspond to
other-determined choice (you get cheese or ice cream at the chef's
discretion).%
\footnote{\small The very idea of defining dual logical operators in
  terms of choice accorded to different players in a two-person proof
  game goes back to C.S. Peirce, who used it informally to
  characterize $\forall$ and $\exists$.  In the 1950s, Paul Lorenzen
  also applied it to the pair of conjunction and disjunction as they
  occur in intuitionistic and classical logic.}
The menu modelling surely serves the cause of logic.  Yet I should
prefer not to postulate, \textit{as a first interpretive step} in
mathematical semantics for the vernacular, a \textit{logical
  ambiguity} behind these phenomena.  (A proposal for `\textit{or}'
similar to Lafont's is Barker 2010.) One reason for being sceptical of
this investigative tactic, even for a language fragment having
`\textit{or}' as its only connective particle, are the very subtle
pragmatic concomitants of the difference between `free choice' and
other readings of `\textit{or}' (see Merin 1992). For a more direct
approach to `\textit{or}' along the lines of Merin (1986), see the
Appendix.

The framework of substructural logics brings to formal fruition a
dream of Carnap's in granting logicians the utmost freedom to develop
tailor-made derivation systems.  At present, I do not see how the
descriptive problem turning on the lattice laws can be solved in this
framework.  Others might succeed in doing so.%
\footnote{\small An application of substructural logic to vernacular
  `\textit{if}', including related uses of `\textit{or}' is Paoli
  (2012), whose logic HL has three kinds of each. See note\:\ref{EXCO}
  below for a paradigm example of the general methodological issue.}
If so, the body of the present article should yet motivate a need for
their endeavours. Its main objective, however, was and is (i) to note
a pervasive empirical problem in the parlour or vestibule of
philosophy, (ii) to indicate how philosophy has managed to ignore it,
and (iii) to affirm that this is an instance of a general
methodological problem.

Suppose the quest for auxiliaries that preserve non-sub-structural
logic proves futile.  And suppose also, perhaps prematurely, that
sub-structural logics, too, do not afford a remedy for a sizeable
fragment of the vernacular. Or suppose they do, but would saddle one
with homophone connectives for which the often alleged and now long
discredited ambiguity of `\textit{or}' between '$\lor$' and
`\textsc{xor}' could be paradigmatic. Are we then left with a return
to a theory-less theory of meaning for the vernacular? Would the only
choice for theory-minded philosophers be one between a leap of faith
in things as they are held to be and the deep blue sea of
nihilism, i.e.~of anti-mathematical philosophy?

I do not think so. However, without a lengthy exposition -- for which
there is no room in this article-sized essay -- the proposal of any
conceivable non-conservative alternative must be a largely unsupported
statement.  Presenting an idea as a statement that lacks detailed
substantiation is the philosopher's equivalent of science fiction. I
literally present the idea as such in the Appendix, because this seems
like the proper register for a three-page memorandum. 

A non-lattice-theoretic algebraic semantics such as the one to be
fictionalized need not dispense altogether with a logic that validates
notably (Ide). We can at least verbally conceive of such an
alternative approach to linguistic meaning as being based on a
reversal of priorities.  Instead of a skeleton of logic wrapped in
pragmatics, language could have a skeleton of pragmatics which, every
now and then, is corseted or even stopped dead in its walkabout tracks
by logic of a most classical kind.  Pragmatics, if very abstractly
conceived, could be as articulate, indeed, in Boole's terminology: as
algebraic, as logic.

Let us not take this for granted. Suppose merely that the badness of
exx. (2b) and (5c,c$'$) is due -- somehow -- to an offense against
(Ide.2).  This, after all, and I speak quite unhypothetically now, is
what it will feel like when you reflect on your aesthetic
apperceptions.  Examples (2b), (5c,c$'$) and (6c) grate on the mind's
ear in much the same way.  If so, our meaning theory should have to
\textit{explain how a meaning is generated that can offend against
  logic in the first place}.  Bare sentence-formation syntax cannot do
this, for it is meaningless by definition.  By definition, logic as we
mostly know it cannot do this either. A logic that did would have to
invalidate, for one, (Ide.2); recall p.\:\pageref{LOGICS}.  Thus,
something else is needed and logic as we mostly know it would only cut
in at some point, quite late in the interpretive day and perhaps in a
sparse way.  In return, it would make its entry with a bang -- here:
Boole's fundamental law coming down hard on perceived irrationality --
rather than cut out with a whimper, as I believe it will have to when
under the influence of Gricean ambitions. Let `logic' once again refer
to logic as most philosophers and working mathematicians know it. Let
non-logical or sub-structurally logical theories of meaning refer to
the relevant complement. If meanings generated from within this
complement conform to the requirements of logic so circumscribed,
there is no way to distinguish between the Gricean approach and a
non-conservative alternative.  But if language is bumping into logic
in broad daylight, it must in the first place be heedless of logic.

\section{Normativity, description, and `reflective equilibrium'}

One might reply: Grice, either in person or \textit{pars pro toto} for
the Gricean enterprise, has saved a logic-based meaning theory in the
past, so he will save it this time too. Or rather: one might think so,
but not say so, and there would be a good reason for discretion. The
thought is not unlike the inductive \linebreak 
reasoning of Russell's chicken
which had its neck wrung by the hand that used to feed it daily. In
fact, the thought's inductive base might be more slender than the
chicken's, to the extent that past Gricean claims turn out to have
been illusory (recall note\:\ref{HAIR}). The chicken was at least fed
real chickenfeed.

All of this sounds so very negative. Let us then think positive.
First, a \linebreak 
denial of the descriptive adequacy of supplemented logic for
paraphrastic equivalence data entails a corresponding \textit{denial
  of the most obvious form of psychologism about logic}.  Our
vernacular language is an object of social psychology.  Were it to
conform at heart to the norms of such-and-such a logic, who could say
that this alleged norm of how we ought to reason is not simply a law
or requirement of our psychology, much as Boyle's law about gases is a
law of physics? Since it does not so conform, philosophers who have
the will to believe in antipsychologism and the normativity of logic,
but who are not yet fully convinced in their heart, now have an extra
plausibility argument to boost their faith.

There are other ways, too, in which the denial of a logical base to
linguistic meaning does not impugn the role of logic in analytic
philosophy.  Informed respect for logic is what distinguishes the
would-be Analytic community most clearly from its Continental
\textit{b\^ete noire}.  This distinctive role of logic is more easily
recognized than that of other branches of mathematics in philosophy,
say, probability or whatever else it takes to do philosophy of
science. What makes logic distinctive \textit{qua} mathematics is that
a logic has a consequence relation -- a specification of what must be
undeniable if such-and-such is affirmed -- which indeed defines it.
And consequence is undeniably at the heart of all philosophical
argument, even if in actual application our notions of consequence may
differ subtly from the idealizations of our preferred logician.  This
distinctive role of logic will continue to be backed up by content
even if the vernacular-generating thesis fails, as I think it does.
Students and users of logic have many more strings to their bow than
this particular application.  Logic and logics as pursued by logicians
in the \textit{Journal of Symbolic Logic} and several more recently
established journals are part of pure mathematics, like geometry and
its plurality of geometries.  Logics have applications for engineering
purposes and for the philosophical reconstructive description of
mathematical and scientific practice.

The lastmentioned, descriptive applications of logic are in many ways
independent of its relation to the vernacular. For example, a very
simple example, there is a way to specify inclusive disjunctions of
actual or potential measurement observations without using the word
`\textit{or}'. We say `at least one of $A$ and $B$ is true'.  A
limited and imperfect fit of logic to the vernacular would suffice to
keep us talking nearly enough in line with the norms of our working
logic.  For conservative analytic philosophers, this logic will by
default be classical logic.%
\footnote{\small \label{EXCO} Use of its dreaded explosive device,
  \textit{ex contradictione quodlibet sequitur} -- put simply:
  \mbox{$A\,\!\land\!\,\neg A$} entails \textit{any} $B$ -- will be
  proscribed if we reasonably require that premisses be evidentially
  relevant to conclusions. When relevance of $X$ to $Y$ is explicated
  in probability theory, i.e.~as $P(XY)\neq P(X)P(Y)$, then
  $A\land\neg A$ is always irrelevant to all $B$ under all $P$.
  This example could serve as a paradigm for
  comparing as instruments for explicating vernacular inferential
  intuitions (i) non-classical logics, among them both paraconsistent,
  i.e.~non-explosive logics and `relevant logics', with (ii)
  classical logic supplemented and thence constrained by classical
  probability theory or all but embedded in it as the logic of the
  underlying proposition algebras.}

A limited fit does not mean that `\textit{and}', `\textit{or}', etc.
`\textit{never mean}' what `$\land$', `$\lor$', etc. mean. Limited fit
would suffice for, and would not rule out, the felicitous
reconstruction of many philosophical and everyday arguments in which
these English words occur by direct translation of the very words into
the familiar logical correlates. In a passing remark dropped in the
most elegant of logic primers, E.J.\:Lemmon (1965:167) surmised that
sentences of our vernacular do not \textit{per se} have logical forms.
Rather, he says, it is arguments in which sentences are used that have
such forms. This way of identifying the home ground of our best known
logics suggests a descriptive alternative to pursue:
\begin{quote}\label{INSET}
  Rather than assume that language is logic in a wrapping of mostly
  parochial syntax and largely universal pragmatics, we could conceive
  of language predicated on alternative forms of meaning composition
  at base. However, in sufficiently many contexts of indicatival use
  and in concert with other constraints, this extra-logical base would
  induce commitments to belief that each conform to the prescriptions
  of our favourite logic.

  A heuristic analogy would be our use of dead metaphor, say, `the
  last leg of the journey'. With dead metaphor we mean one thing,
  without any metaphoric stretching felt, by means of an expression
  whose literal, that is, compositional meaning is something other.
  That meaning is presumed dead, but may turn out to be undead, like
  Count Dracula at night-time, in certain contexts of use or on being
  tweaked by the punster.
\end{quote}
Autonomous logic and mathematical or scientific practice would take
over where the vernacular fails to coincide with the requirements of
practice as codified in a logic. Arguing about which logic is right, or
right for which purpose, is arguing about what it means to be
rational.%
\footnote{\small I am pretending that attention can be confined to deductive
  logic. In actual fact, it must extend to probability or other
  frameworks for reasoning under uncertainty.}

We have to live with the profoundly emotive term `rational', I guess.
What we should not take for granted, though, is appeal to the kindred
term \linebreak `reflective equilibrium'. This mellifluous expression
will suggest, and perhaps indeed refer to, the terminal state $\sigma$
of an iterative procedure $F$ of reflection which remains stable under
more reflection, $F(\sigma)=\sigma$, and so affords both theoretical
perfection and of tranquility of mind.  However, I believe that there are
referents of greater argumentative importance and that the relevance
of the headline referent is mainly to lend their use more
\textit{gravitas} and goodness.

The process of reflection could be pictured as a dialectic among
multiple \textit{mentis personae} of the reasoner, call them
Face-the-facts, Give-us-norms and Least-effort. `Equilibrium' could
then refer to its standard game-theoretic instance: a combination of
choices by all players (each player choosing one among his options for
individual action) that jointly determines their individual payoffs
and such that no player can improve his position by a unilateral
change of choice. With these \textit{personae}, equilibrium combines
nicely with physiomorph images of an equilibrium of forces or with
sociomorph images of equitable division.

However, in games there need be nothing globally optimal let alone
fair about an equilibrium. Being stuck in a suboptimal if not pessimal
equilibrium is a salient predicament in interactive decisionmaking.
And knowing our three players, the game will be one of divide-the-pie,
and the favoured equilibrium most likely one where Give-us-norms and
Least-effort divide the pie among themselves. This predicament is
indeed what I believe the most important current use of the phrase is
apt to get the philosopher into.

There are good cases of it, when the recommendation is that we
regiment our professional usage. We conduct our arguments in a
language of logic that had its functional vocabulary `syntactically
sugared' to resemble English, as computer scientists would say, and we
are out-front about this. The bad cases arise when the provenance of this
language is forgotten and when appeal to Mr and Ms Natural's
vernacular language intuitions is made in philosophical argument about
what language (or mind) is.  It is in this grey zone of equivocation%
\footnote{\small Here is a poetic instance from neighbouring linguistics.
  Having exemplified (Dis.1) (p.\:\pageref{DIST} above) in English,
  Keenan and Faltz (1985:71) invite their readers to ``construct an
  example showing that [(Dis.2)] should be satisfied''.}
that appeals to reflective equilibrium or a tacit `don't look now'
allow philosophy to have its cake and eat it. Some reasons have been
outlined in the first section why an appeal to reflective equilibrium
will also sustain peace of mind of the not-for-profit variety.
However, to understand peace of mind as such it helps to consider its
opposite: disquiet. Here is some exploratory fieldwork on it.

A philosophical logician with a keen interest in philosophical thought
experiments was asked, in 2012, to consider hypothetically the
following question: What would it feel like if it turned out that our
vernacular is not based on a skeleton of logic in which `\textit{and}'
means $\land$ and `\textit{or}' means $\lor$? His reply was that it
might feel as if `Seven plus five is twelve' turned out not to mean
`$7+5=12$'.  I should add my own bit to this \textit{impromptu}
intuition, in line with the inset proposal on p.\:\pageref{INSET}
above. On present showing, I feel that the two expressions would not
mean the same when considered compositionally. Yet any utterance of
the first sentence would, I also feel, continue to mean what
`$7+5=12$' means.  The combination of these two properties would feel
as if we did not quite know what we are saying, and in a sense much
more acute than might be claimed for dead metaphors which we are
sometimes said to live by.

When it comes to our own language, there are good reasons, then, to
wish to be able to believe that logic is descriptive, too. The
received view on this has a counterpart in physics. We believe in
Newtonian rigid body mechanics, the mechanics of conservative forces.
We believe in it in spite of trolleys slowing down without an extra
push or pull and in spite of feathers falling more slowly than
pebbles.  We do so, because an auxiliary theory of friction, i.e.~of
non-conservative forces that turn kinetic energy into heat, and of
aerodynamics is available to us. In reflecting on earthbound
mechanical engineering purposes that can take materials for granted we
do not have to worry either whether classical, Newtonian mechanics is
indeed a special case of relativistic mechanics or how it can
articulate with quantum mechanics.  Gricean and perhaps post-Gricean
supplements play the role of a classical auxiliary for logic (as
commpnly understood)..

But suppose we come across phenomena for which there is no respectable
auxiliary theory in sight.  In such moments, appeals to `reflective
equilibrium' tend to be made; and in such moments, philosophy begins
to differ decisively from physics and the other natural sciences.
Should it not differ from them anyway?  No doubt it must, but I do not
see why it should differ on this point of method.  Philosophy, when it
makes claims about -- not simply claims on -- the vernacular language,
is, after all, making empirical claims.  It cannot all farm them out
to linguistics, for if it did, we should have to stop doing and 
teaching philosophy of language and much of philosophical logic.

Appeal to reflective equilibrium under which phenomenology and its
observables must give way to a coalition of prescription and economy  
of thought obscures the taxonomic fact that systematic
philosophy is in parts an empirical discipline.  Philosophy can steer
clear of such appeals if it takes care to distinguish its normative
and its descriptive aspects and to keep each one from subverting the
core business of the other.  There may be areas of\linebreak
philosophy in which it is difficult to make a workable distinction,
but the \mbox{theory} of meaning is not one of them.

\small
\newpage
\section*{Appendix: 
The View from Triple Sec}

Triple Sec is a planet of Beta Chimerae, one orbit outward from Twin
Earth. The Triple Sec Institute of Philosophy [TSIP] have made a study
of English, presently the most natural language on Twin Earth.  They
have hit on the idea -- congenial to their conservationist mindset -- that
speakers of \mbox{English} sentences are, in the first place, both
imaginative-intuitive beings (\textit{Anschauungswesen}, as their Kant
scholars\linebreak 
germanize) and passionate-desiderative beings.  Accordingly,
says a TSIP spokesperson, sentences of English can be expected to have
their natural interpretations of first resort not in boolean or
similar lattice algebras of truth or proof conditions, but in rather
different mathematical structures. These have for instances, on the
one hand, the Euclidean spaces of geometry, physics, and statistics,
and, on the other hand, the commodity and service bundle spaces of
economics.  Instances of the first kind also include spaces of
representations by images, as familiar from handmade and computer
graphics. Images are not truth- or proof-valued by constitution. The
economically interpreted spaces are likewise structured, not by truth
and consequence, but by comparative and quantitative preferences, i.e.
by essentially pragmatic value relations. Their objects are
preference-valuables and disvaluables.  If objects from either kind of
space were meanings of sentences, they would, by definition, be
non-propositional meanings, at any rate to start with.

What all these spaces have in common, so TSIP scholars now observe, is
that they are linear algebras -- of the most familiar kind predicated
on the intuitive notion of \textit{quantity}, i.e.~over ordered rings
or fields, as mathematicians say, and thus nothing exotic based on
number systems in which $1+1=0$ or $7+5=1$. Linear algebras are also
known as vector spaces.  (TSIA operatives on Twin Earth report that
their rudimentary doctrine, initiated by one Des Cartes, is taught
there at pre-university stage; and the core of it, under cover of
`arithmetic', already to six-year-olds!)  Twin Earth English
sentences, say, $A$, will thus denote abstract objects, $\mathsf{A}$,
called `vectors', just as English sentences on remote Earth are said
by colleagues there to denote abstract objects which are elements of
boolean algebras and are called `propositions'.  And just as
`proposition' suggests a certain argumentative interpretation, so TSIP
scholars, punning on vistas and economic expectations, call Twin Earth
sentence meanings `prospects'. Prospects, including those denoted by
connective-free sentences are, in turn, componible from phrasal and
content-word meanings that are also elements of linear spaces and
combine as suitably dimensioned vectors and linear maps or, in
suitable circumstances, tensor product formation. TSIP methodologists
have noted that structural `distributional semantics' for content
words, elicited in computational linguistics by statistical latent
structure analysis, also finds meanings in linear spaces; but makes
for a rapid exit to truth-valued propositions when it comes to
sentence meanings (so in Coecke, Sadrzadeh \& Clark 2010).  TSIP
scholars, traumatized by data on absorption, distribution etc., prefer
to give prospects a ride for their money before eventually and 
gingerly relating them to propositions.

Sentence-conjoining `\textit{and}' will thus denote vector addition,
`$+$', of prospects; analogous things will hold for phrasal
conjunction.  (TSIP investigators find unprofitable for their
immediate purposes a `quantum logic' use of linear spaces, which by
way of denotations in subspace lattices leads straight to `meet' and
`join' connectives; see note\:\ref{QUANTUM}.)  The laws of vector
addition closely resemble those of arithmetical addition, but vectors
need not be numbers, so visualize vector `$+$' as `$\oplus$', if that
liberates the imagination.  The key algebraic difference between `$+$'
and logical `$\land$' is that $\mathsf{X}+\mathsf{X}=\mathsf{X}$ is
invalid.  Vector addition $\mathsf{X}+\mathsf{Y}$ is generalized in
linear algebra to `linear combination'
\mbox{$a\mathsf{X}+b\mathsf{Y}$} of vectors $\mathsf{X}$ and
$\mathsf{Y}$, where $a$ and $b$ are real number coefficients, called
`scalars'.  Addition is the special case $a=b=1$. TSIP poetologists
reply to worries about $a=-500$ and $b=\sqrt{2}$ that a semantics for
a language $L$ in a combinatorially generable or other domain of
interpretation $D$ must attach to every sentence of $L$ an object in
$D$, but need not require every object in $D$ to be expressible in
$L$. Real junk, they say, will be unspeakable. For \mbox{`\textit{X or
    Y}'}, they hypothesize context-indexical, scalar-valued
coefficient variables constrained by \mbox{$\{0,1\}\ni b=1-a$} and
thus $a\mathsf{X}+(1-a)\mathsf{Y}$.  The choice of value is
\textit{ceteris paribus} unspecified by verbal means, but is by
indicatival convention left to \mbox{Nature}, whose choice a cagey
speaker may be privy to or even execute; or, in deeply imperatival
discourse, to the addressee.  Each occurence of `\textit{or}' gets a
\textit{prima facie} independent choice and thence a distinct
coefficient variable.

Despite superficial appearances, Twin Earth `\textit{or}' is not
logical \textsc{xor}.  Inclusive and properly exclusive readings
require induction by material or rhetorical interests that are
imputable in a given context of use.  The TSIP bargain basement of
Twin Earth ethnographica has a hint on offer: `\textit{You may take my
  wallet or my bike}' tends to be read exlusively, or rather: `at most
one of them'.  By contrast, `\textit{You must give me your wallet or
  your bike}' reads dually, `at least one of them' and thus
inclusively.  TSIP management conclude that competent speakers of Twin
Earth English appear to be veritable \textit{homines oeconomici}.

TSIP advice to mental space travellers who would dream of mapping
`\textit{or}' on two or more distinct connectives of a substructural
logic such as Linear Logic is, accordingly, pragmatic. Such travellers
had better plan a route by way of a resting place: ordered linear
spaces and a single, univocal, if intrinsically indexical connective
operation in the linear combination family.  Against that backdrop,
which sets a minimum standard of descriptive adequacy, they might
profitably investigate for their descriptive potential pure
substructural logics, say, logics with a constant-sum game semantics.
(Andreas Blass, so the TSIA residency on remote Earth tells them, has
offered such semantics for linear logic, as presaged by a fundamental
connection to abstract games spelt out early on by Yves Lafont.)
 After
this excursion into what TSIP scholars still consider science fiction,
let us return to their perceived reality.

Assuming that `\textit{A and A}' designates $\mathsf{A}+\mathsf{A}$,
it can mean $\mathsf{A}$ only if `\textit{A}' means \mbox{Nothing},
the null vector.  For iterables (e.g.~`\textit{Kim talks and (Kim)
  talks}') additivity is just fine. For stative `\textit{A}' (e.g.
`\textit{Kim is tall}') it generates what TSIP observers call a
`double image'. It offends the Twin Earth ethics of thought for sober
argumentation -- also subscribed to on Triple Sec -- which is
enshrined, for one, in (Ide.2).  By contrast, \mbox{`\textit{A or
    A}'}, designating $a\mathsf{A}+(1-a)\mathsf{A}$, will always
denote $\mathsf{A}$, i.e.~ what `\textit{A}' denotes, regardless of
whether $a$ is instantiated to 0 or to 1. Its oddity, say TSIP
contemplators, arises solely from its being an instance of Hobson's
Choice. Now, linear combination obeys in all essentials the
distributive law $(x+y)z = xz+yz$ of plain arithmetic.  The TSIP
report accordingly observes that the offending, fully inacceptable
examples (2b, 5c) all have options, for some possible assignments of 0
and 1 to scalar variables in occurrences of `\textit{or}', where a
double image appears among the possible options.

Thus, $\mathsf{A}+[a\mathsf{A}+(1-a)\mathsf{B}]$, of which (5c) is an
instance, equals $\mathsf{A}+\mathsf{A}$ when $a=1$ and it equals
$\mathsf{A}+\mathsf{B}$ when $a=0$. The first option is a double
image.  It cannot simply be ignored: doing so would make `\textit{A,
  and A or B}' equivalent to `\textit{A and B}'. Neither can logic cut
down the first option to sensible $\mathsf{A}$ without a fuss.  If it
did, `\textit{A, and A or B}' would intuitably (not just normatively)
mean what `\textit{A, or A and B}' means, since
$a\mathsf{A}+(1-a)[\mathsf{A}+\mathsf{B}]$ reduces to $\mathsf{A}$
when $a=1$ and to $\mathsf{A}+\mathsf{B}$ when $a=0$.  The form
`\textit{A, or A and B}' is perfectly acceptable and its
interpretation, so the TSIP report surmises, is correspondingly
intuitive.  By similar computations, and remembering to use distinct
coefficients for distinct occurrences of `\textit{or}', TSIP scholars
have also verified that (1a) and (1b) denote identical sets of vector
options, while (2a) and (2b) do not. TSIP refers to Merin (1986, 1997,
2012) for considerations on negation, on probabilistic evidential
relevance linking linear prospects and boolean propositions, and on
predicate languages with multilinear semantics which allow people to
say and mean things like `\textit{Kim and Lee sang or danced}' or
`\textit{You and you owe me a drink}'.


\vspace{0.5cm}

\newpage
\normalsize
\section*{References}
\begin{description}
\item 
Adams, E.W. (1975). \textit{The Logic of Conditionals}. Dordrecht: Reidel.
\item\vspace{-10pt} 
Barker, C. (2010). Free choice permission as resource-sensitive reasoning.
\textit{Semantics and Pragmatics} 3, 10:1-38.
\item\vspace{-10pt} 
Birkhoff, G. \& von Neumann, J. (1936). The logic of quantum mechanics.
\textit{Annals of Mathematics} 37, 823--843.
\item\vspace{-10pt} 
Black, M. (1952). Saying and disbelieving. \textit{Analysis} 13, 25--33. 
\item\vspace{-10pt} 
Boole, G. (1854).
\textit{An Investigation of The Laws of Thought on which are founded the 
Mathematical Theories of Logic and Probabilities}.
London: Macmillan. Repr. New York: Dover 1958.
\item\vspace{-10pt} 
Coecke, B., Sadrzadeh, M. \& Clark, S. (2010). Mathematical 
foundations for a compositional distributional
  model of meaning. arXiv:1003.4394 [cs.CL]  [34 pp.]
\item\vspace{-10pt}  
Felgner, U. (2002). Editor's notes to F. Hausdorff 
\textit{Grundz\"uge der Mengenlehre} [1914] repr. Berlin: Springer. 
\item\vspace{-10pt}  
Frege, G. (1879). \textit{Begriffsschrift}. Halle: Louis Nebert. 
[Trsl. by S. Bauer-Mengelberg in van Heijenoort (ed.), 5--82.] 
\item\vspace{-10pt} 
Gazdar, G. (1979). \textit{Pragmatics: Implicature, 
 Presupposition, and Logical Form}. London: Academic Press.
\item\vspace{-10pt} 
Gentzen, G. (1934). 
Untersuchungen \"uber das logische Schließen.
\textit{Mathematische Zeitschrift} 39, 176--210, 405--431.
\item\vspace{-10pt} 
Girard, J.Y. (1987). Linear logic. 
\textit{Theoretical Computer Science} 50, 1--102. 
\item\vspace{-10pt} 
------. (1990). La logique lin\'eaire. \textit{Pour la 
Science} 150, 74--85. 
\item\vspace{-10pt} 
Grice, H.P.  (1961). The causal theory of
  perception. \textit{Aristotelian Society Suppl. Volume} 35, 121--152.
  Partially repr. in Grice (1989) [the part about `or' is omitted].
\item\vspace{-10pt}  
------ (1967). Logic and Conversation. Ms. Harvard University. 
Repr. with revisions 
in Grice (1989).
\item\vspace{-10pt}  
 ------ (1989). \textit{Studies in the Way of Words}.
 Cambridge MA: Harvard University Press.
\item\vspace{-10pt} 
Keenan, E. \&  Faltz, L (1985). \textit{Boolean Semantics for
   Natural Language}. Dordrecht: Reidel.
\item\vspace{-10pt} 
K\"onig, J. (1914). \textit{Neue Grundlagen der Logik, 
Arithmetik und Mengenlehre}. Leipzig: Veit \& Compagnie. 
\item\vspace{-10pt} 
Lambek, J. (1958). The mathematics of sentence structure.
\textit{American Mathematical Monthly} 35, 143--164.
\item\vspace{-10pt} 
Lemmon, E.J. (1965). \textit{Beginning Logic}. London: Nelson.
\item\vspace{-10pt} 
Lewis, D.K. (1973). \textit{Counterfactuals}. Oxford: Blackwell. 
\item\vspace{-10pt}
McCawley, J.D. (1981). \textit{Everything that linguists have 
always wanted to know about logic (but were afraid to ask)}. 
Chicago: University of Chicago Press. 2nd edn. 1993.
\item\vspace{-10pt}
Mendelson, E. (1970). \textit{Boolean Algebra and Switching Circuits}. 
New York: McGraw-Hill.
\item\vspace{-10pt} 
Merin, A. (1986). `Or', `and': non-boolean utility-functional connectives.
 \linebreak 
$[$Abstract$]$ \textit{Journal of Symbolic Logic} 51, 850--851.
\item\vspace{-10pt}  
------ (1992). Permission sentences stand in the way of 
        Boolean and other
        lattice-theoretic semantices.  \textit{Journal of Semantics} 9, 
95--162.
\item\vspace{-10pt} 
------ (1994). \textit{Decision-Theoretic Pragmatics}. Lecture Notes
European Summer School in Logic, Language and Information (ESSLLI'94).
Copenhagen: Copenhagen Business School.
\item\vspace{-10pt} 
------ (1997). If all our arguments had to be conclusive, there would be
few of them. \textit{Arbeitsberichte des SFB 340}
Nr. 101, Universities of Stuttgart and T\"ubingen. Online at\\ 
http://www.ims.uni-stuttgart.de/projekte/SFB340.html and as\\
$\langle$http://semanticsarchive.net/Archive/jVkZDI3M/101.pdf$\rangle$.
\item\vspace{-10pt} 
------ (1999). Information, relevance, and social
  decision-making: some principles and results of Decision-Theoretic
  Semantics.  In: L.S. Moss, \linebreak J. Ginzburg \& M. de Rijke
  eds.  \textit{Logic, Language, and Computation} Vol.\:2. Stanford CA:
  CSLI Publications, 179--221.\\
  Online: $\langle$http://www.let.uu.nl/esslli/Courses/merin/irsdmu.ps$\rangle$
\item\vspace{-10pt}
------ (2006). L'anaphore des ind\'efinis et la pertinence des
  pr\'edicats. In
 F. Corblin, S. Ferrando and L. Kupferman (eds.) \textit{Ind\'efini 
et pr\'edication}.
  Paris: Presses Universitaires de Paris-Sorbonne, pp.~535--550.\\
(online: 
http://semanticsarchive.net/Archive/DQyZmNhM/adipp.pdf\:)
\item\vspace{-10pt} 
 ------ (2012). Multilinear Semantics for Double-Jointed and Convex 
 Coordinate Constructions. $\langle$http://semanticsarchive.net/Archive/mJjNTIwY/
\linebreak
MultilinearSDJCCC-Merin.pdf$\rangle$
\item\vspace{-10pt} 
O'Hair, S.G. (1969). Implications and meaning. \textit{Theoria} 35, 38--54.
\item\vspace{-10pt} 
Quine, W.V.O. (1950). \textit{Methods of Logic}. New York: Holt. 
(British edition: London: Routledge and Kegan Paul, 1952.)
\item\vspace{-10pt} 
Paoli, F. (2002). \textit{Substructural Logics: A Primer}.
Dordrecht: Kluwer. 
\item\vspace{-10pt} 
------ (2012). A paraconsistent and substructural
  conditional logic. In K. Tanaka et al. (eds.)  
\textit{Paraconsistency: Logic and
    Applications}.  Dordrecht: Springer, Ch.\:11., pp. $x$--$x$+25.
\item\vspace{-10pt} 
Restall, G. (2000). \textit{An Introduction to Substructural Logics}.
London: Routledge.
\item\vspace{-10pt} 
Schr\"oder, E. (1890).
  \textit{Algebra der Logik}. Vol. I
  Leipzig.
  Repr. Bronx NY: Chelsea Publishing Company, n.d.
\item\vspace{-10pt} 
Soames, S. (1982). How presuppositions are
  inherited: a solution to the projection problem. \textit{Linguistic
  Inquiry} 13, 483--545.
\item\vspace{-10pt} 
 Tarski, A. (1946). \textit{Introduction to Logic and to the 
Methodology of the Formal Sciences}. 2nd edn. New York: Oxford 
University Press. 
\item\vspace{-10pt} 
van Heijenoort, J. [ed.] (1967). \textit{From Frege
    to G\"odel: a Source Book in Mathematical Logic, 1879--1931},
  Cambridge MA: Harvard University Press.
\end{description}


\vspace{-0.5cm}

\footnotesize\hbox{}
\vspace{-3pt} 
\hbox{}
\noindent
Author's electronic address:\\
\vspace{-3pt}
\noindent
\hbox{}\hspace{-3pt}arthur.merin@uni-konstanz.de
\normalsize

\end{document}